# Designing a Trusted Data Brokerage Framework in the Aviation Domain

Evmorfia Biliri[1], Minas Pertselakis[1], Marios Phinikettos[1], Marios Zacharias[2],
Fenareti Lampathaki[1], Dimitrios Alexandrou[3]

[1] Suite5 Data Intelligence Solutions Limited, 95B Arch. Makariou III
3020, Limassol, Cyprus
{evmorfia, minas, marios, fenareti}@suite5.eu

[2] Singularlogic Anonymi Etaireia Pliroforiakon Systimaton Kai Efarmogon Pliroforikis, 3
Achaias 145 64, Kifisia, Greece
mzacharias@singularlogic.eu

[3] UBITECH, 8 Thessalias 152 31, Chalandri, Greece
dalexandrou@ubitech.eu

**Abstract.** In recent years, there is growing interest in the ways the European aviation industry can leverage the multi-source data fusion towards augmented domain intelligence. However, privacy, legal and organisational policies together with technical limitations, hinder data sharing and, thus, its benefits. The current paper presents the ICARUS data policy and assets brokerage framework, which aims to (a) formalise the data attributes and qualities that affect how aviation data assets can be shared and handled subsequently to their acquisition, including licenses, IPR, characterisation of sensitivity and privacy risks, and (b) enable the creation of machine-processable data contracts for the aviation industry. This involves expressing contractual terms pertaining to data trading agreements into a machine-processable language and supporting the diverse interactions among stakeholders in aviation data sharing scenarios through a trusted and robust system based on the Ethereum platform.

**Keywords: data brokerage, collaboration platform, aviation**

## 1  Introduction

The aviation industry encompasses all the activities that are directly dependent on transporting people and goods by air. This covers airport and airlines operations, aircraft construction and maintenance, air traffic control and regulation, passenger and freight services, among others. Aviation stakeholders produce and consume, nowadays, vast amounts of data and the industry is already investing on them, aiming to utilise their full potential in order to improve passenger experience and flight efficiency, expand sales and reduce costs.

In this direction, the emergence of big-data technologies has pushed the aviation domain many steps forward: the collection and storage of massive data sets has become easier, the parallel processing provides the necessary computation infrastructure, and data science has developed the prerequisite tools that can provide insights and predictions directly from high volumes of data. Recent academic works on machine learning techniques for aviation applications provide strong proof regarding the ways in which data analytics can contribute to significant domain issues, indicatively including air travel demand modelling and operational safety and quality [1], [2], [3]. However, they also hint to the lack of cross-section data availability in real-world cases, i.e. at scale, which hinders the development of more powerful multi-source data analytics solutions. One of the main reasons for this is that the highly competitive business players of the aviation sector remain sceptical when data diffusion and sharing comes into the discussion. Beyond technical limitations that need to be overcome, establishing trust and fairness in the scope of data sharing is also an important, yet highly challenging first step towards a sustainable collaborative network spanning across the broader aviation ecosystem.

Several EU projects and works in literature have tackled in the past the challenges of collaborative networks and platforms for enterprises [4],[5] even for data sharing [6]. Most of these solutions however prefer to follow a more generic approach, ignoring the special characteristics and peculiarities of serving a particular domain. The few exceptions to this rule, like [7], which explores business scenarios in the solar energy domain, do not involve the data analytics target of collaboration, nor the data sharing and asset exchange using a clear and secure framework for intellectual property rights (IPR).

The ICARUS project aims to fill this gap by bringing together all aviation related stakeholders and accelerate their collaboration on data exploration and analysis through an innovative big data enabled sharing and collaboration platform, removing current barriers in data aggregation, sharing and IPR protection. These features, combined with a targeted aviation oriented data model and metadata model, constitute the novelty of the proposed framework.

## 2   Background

### 2.1  Data Sharing Motivation and Initiatives in Aviation

Data sharing is not a new concept for the aviation industry. GAIN, the Global Aviation Information Network, was proposed by the Federal Aviation Administration (FAA) in 1996. Since then, numerous multi-airline and multi-national data sharing programs and initiatives which involve centralising airline flight data storage have been established: Flight Data eXchange (FDX) is an aggregated de-identified database of FDA/ FOQA type events that allows to identify commercial flight safety issues for a wide variety of safety topics. The ASIAS program, developed by FAA and the aviation industry aims to promote an open exchange of safety information. Over 90% of IATA member carriers have agreed to participate in GADM, the IATA Global Aviation Data Management programme and platform. STEADES$^{TM}$, the IATA's aviation safety incident data management and analysis program that constitutes one of the GADM data sources, has over 200 members. SKYbrary is an electronic repository of safety knowledge related to flight operations, air traffic management (ATM) and aviation safety. Partners of the European Airport Collaborative Decision Making (A-CDM) share timely information through adapted procedures and tools enabling real-time collaborative decision-making.

The information being shared is broad and includes, among others, aeronautical data, flight trajectories, aerodrome operations, historical and current meteorological data, surveillance data (e.g. from radars). The primary goal of such initiatives is to ensure safety in the air travel, which in turn requires the optimisation of a wide range of operations, e.g. the way Airline Operations Centres plan flight routings. Nevertheless, the launch of such Aviation Data Exchange programmes has opened the door to data sharing with trusted third parties [8].

Advantages of data sharing in the aviation industry are numerous and multi-facetted and bolster the development of further larger-scale collaborations. IATA, Eurocontrol and other core stakeholders of the aviation industry but also stakeholders of the broader spectrum, all report on the expected advantages of collaborations built on shared data and insights [9]. Ad-hoc collaborations, such as EasyJet's partnership with Gatwick Airport [10] and Aer Lingus' partnership with the Dublin airport [11], are emerging, while Airbus has launched the Skywise platform that aspires to become the reference platform for core aviation stakeholders. However, an inclusive solution for the aviation industry has not been established yet.

### 2.2 Data IPR and Marketplaces

Even when incentives for data sharing are strong and the expected benefits are well comprehended, privacy, legal and organisational policies and even infrastructure limitations may hinder data sharing and, thus, the benefits that stem from it. An important milestone towards addressing such limitations is the definition of a clear and robust IPR framework. [12] analysed data sharing agreements from industry, academia and government and identified six high-level aspects of data licenses that affect data sharing: (i) attributes regarding the project and the agreement itself, e.g. description of data, (ii) privacy & protection of sensitive information, (iii) access policies, (iv) legal and financial responsibility, data ownership and rights, (v) compliance, (vi) permissible interactions during data handling. Each of the aforementioned aspects encapsulates numerous more specific attributes and properties of the sharing agreement and the underlying data assets. Depending on the context where they are used, specific practices and derived terms are found, e.g. the work presented in [13] regarding only the pricing aspects. The definition of a concrete data IPR handling process to cover broader and more generic data sharing needs, like the ones manifesting in the emerging data marketplaces, is a challenging task, especially when many-to-many data marketplaces are examined, which is the case in ICARUS. As explained in [14], these marketplaces often emphasise on data discoverability and other facilitation activities, including online payment. In their most common form, the platforms do not have ownership of the data, but only act as intermediaries that facilitate transactions, therefore it is extremely important for the interacting parties to be presented with an intuitive yet trustful way of engaging into such transactions which are gradually starting to pertain to machine processable data contracts. In this context, there is an emerging need to develop contract engines that provide querying and validation mechanisms for access to and usage rights of data assets, as well as the status of the agreements being performed. However, the majority of existing marketplaces are far from this level of automation and lack an adequately expressive, but not prohibitively -for implementation- complex, information model [15]. Establishing rigorous provenance through verifiable information for the data being shared/ sold is under this prism of paramount importance to increase trust between interacting stakeholders. Towards this goal, distributed ledger technologies (DLTs) are now being leveraged in the design of the decentralised multilateral platforms (e.g. [16], [17]). The advantages of DLTs in this context are numerous and widely accepted and include transparency and data democratisation [18].

## 3   ICARUS Data Policy and Assets Brokerage Framework

The landscape review presented in Section 2 reveals that although both research- and industry-oriented approaches are emerging for facilitated data sharing in various domains and stakeholders in the aviation industry have begun to grasp the underlying potential of collaborative data analysis, no concrete solution has been proposed and implemented for the aviation industry. One of the key identified reasons is that data trading is not a primary activity for most ATM stakeholders, hence their incentivisation to participate in this activity is largely affected not only by the envisioned benefits, but also by the effort that will need to be devoted into understanding, learning how to use and ultimately engaging a data sharing and collaboration platform. Furthermore, strong KYC (Know-Your-Customer) requirements in the aviation industry, make it hard for stakeholders to trust a broader data marketplace initiative. Finally, apart from the technical difficulties, data privacy and sensitivity aspects especially in the GDPR context are troublesome for many aviation stakeholders.

In this perspective, the proposed ICARUS Data Policy and Assets Brokerage Framework has a dual role:

1. To formalise all data attributes and qualities that affect, or are in any way relevant to, the ways in which data assets can be shared / traded and handled subsequently to their acquisition. This involves licenses, IPR, characterisation of sensitivity and privacy risk levels, but also more generic metadata regarding data content and structure, as well as properties derived from the underlying data model. It should be noted that at least in its initial version, the ICARUS platform will not handle any type of personal data, hence GDPR compliance is not discussed in the current work.
2. To enable the creation of structured, machine-processable data contracts for the aviation industry. This entails the expression of contractual terms pertaining to data trading agreements into an appropriate machine-processable language. The framework foresees the interactions of stakeholders in aviation data sharing scenarios and defines the system's functionalities in this context.

### 3.1 Data Sharing and IPR model

ICARUS adopts a DLT-based solution for data brokerage to ensure the robustness of the framework and increase transparency and trust. The first step towards delivering the ICARUS data sharing system was to select a set of clearly defined properties that can be used to compose a virtual representation of a data sharing contract, i.e. to identify the data license and IPR attributes and policies. For this process, the insights gained through the literature review were combined with findings from the traditional data sharing agreement documents used by the ICARUS industry partners. The ICARUS data sharing framework is built on top of three core entities, namely the Data Asset, the Policy and the Contract and two supporting entities, namely Attributes and Terms, with the latter being specified as one of Prohibition, Permission and Obligation, as shown in Fig.1.

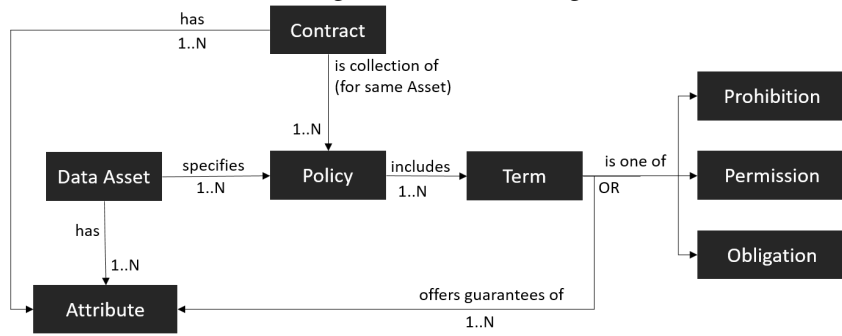

**Fig. 1.** High-level view of ICARUS data sharing model

For each of the presented entities, concrete instantiations are defined, like the ones presented in Table 1 for policies. Balancing expressivity with applicability was necessary, therefore the adopted model is simplified when compared to the complete set of considerations and options of data trading in aviation. Yet the performed assumptions were deemed reasonable for the initial piloting applications of the system and in any case do not harm the future extension and refinement of the framework.

**Table 1.** Indicative terms definition for the ICARUS data brokerage framework

| Policy Category | Indicative Terms |
|---|---|
| Data Asset | description; encrypted & unencrypted columns; included data model entities from the ICARUS aviation data model; provider; creator; contributor; version; created date; modified date; published date |
| Contract | temporal validity; spatial validity & coverage; validation date; liability; involved provider; involved consumer; termination clause |
| Responsibility | ownership; addressed to; liability & indemnification |
| Rights and Usage | license & copyright notice; derivation; attribution; reproduction; distribution; target purpose; target industry; re-context allowed |

| Quality | accuracy; completeness; consistency; credibility; accessibility & online availability |
|---|---|
| Privacy and Protection | privacy & sensitivity compliance (levels, disclaimers, guarantees); liability; applicable law |

### 3.2 Data License and Agreement Manager

The ICARUS data brokerage framework includes the conceptual work on the IPR and license terms, the design of the envisioned and allowed data sharing workflows and, finally, the instantiation of the above in a prototype entitled *data license and agreement manager* that implements the designed functionalities. This component as part of the complete ICARUS platform, is responsible for handling all processes related to the data licenses and IPR attributes, as well as the drafting, signing, and enforcing the smart data contracts that correspond to data sharing agreements between platform users. Its smart contract functionalities are developed on Ethereum, a popular decentralised platform for smart contracts, using the Truffle Framework. The component has three interconnected roles:

1. It allows the users to define, review and update the data license and IPR attributes discussed in the previous section.
2. It allows users to draft, review, negotiate on, and sign a smart data contract that concretely defines the terms under which a dataset will be shared.
3. It handles all processes required to prepare a smart contract for each (paid) asset transaction and, finally, upload it to the blockchain. Depending on the exact terms and attributes included in a contract, the corresponding key-value pairs are stored either as-is or hashed in the blockchain. The component enables the activation (i.e. status change) of a smart contract when both parties - data owner (seller), data consumer (buyer) - approve it and the payment is completed and can also report on the validity of a given smart contract.

Assuming the first step (step 0) of defining the data license, IPR and access policies foreseen by the framework has been completed for a given data asset by its owner (provider), the data brokerage process has the following three high-level phases:

Phase I - Data Assets Exploration: The workflow is initiated by an ICARUS user performing a query to search for data. The search functionality is facilitated by the fact that all data assets conform to the same data model and there is rich additional information defined by the designed metadata and sharing model, which ensures that the results presented to the user adhere to the limitations imposed by the defined terms and policies.

Phase II - Smart Contract Drafting: The step includes the definition of and negotiation upon the terms of the data sharing contract to be signed (Fig.2). Different statuses are foreseen for the smart contract in order to denote whether it is in draft, negotiating, accepted or rejected state.

Phase III - Smart Contract Validation: If the contract reaches the "Accepted" status and the payment obligations are satisfied, the ICARUS platform will allow the consumer to obtain the dataset to use either externally to the platform or to explore and perform analytics on it leveraging the ICARUS functionalities. To avoid issues caused by potentially malicious data providers, a security mechanism to bypass the data provider's validation process is foreseen to address cases where the consumer has proof of a completed payment, but the provider does not honour the agreement.

The proposed framework has been presented to 15 different aviation stakeholders (representing airlines, airports, ground handlers, and generally aviation data providers) during the external validation activities performed in the ICARUS project so far, and has gathered unanimously positive and enthusiastic feedback.

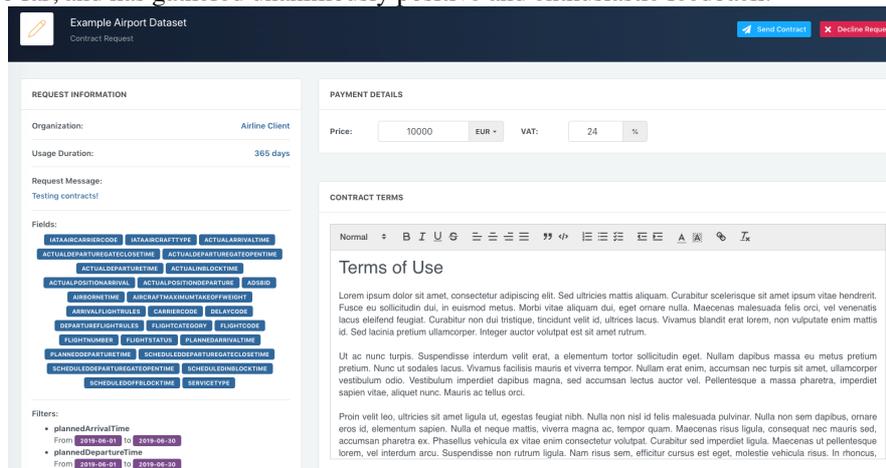

**Fig. 2.** Contract Drafting Page

## 4 Conclusions and Next Steps

The ICARUS policy and brokerage framework sets the foundations of the ICARUS platform that will link data providers and data consumers at all levels of the data value chain in the aviation industry, through a combination of state-of-the-art approaches in academia and industry regarding data brokerage and IPR.

It needs to be noted that the enforceability aspects of smart contracts constitute one of the major challenges, both from a technical and legal perspective. In this context, the future work along the proposed framework will focus on two parallel streams: (a) to investigate how the complexities of multi-sharing, i.e. sharing multiple datasets among multiple stakeholders, can be addressed based on license compatibility analysis, and (b) to conclude the framework's evaluation and validation process by a broader group of aviation stakeholders, which is currently ongoing in the context of the ICARUS project, and leverage the respective feedback to further improve the presented approach.

**Acknowledgments.** This work has been created in the context of the ICARUS project, that has received funding from the European Union's Horizon 2020 research and innovation programme under grant agreement No. 780792.